%% file: neurips_2025.tex
\documentclass{article}

 \usepackage[preprint]{neurips_2025}

\usepackage[utf8]{inputenc} 
\usepackage[T1]{fontenc}    
\usepackage{hyperref}       
\usepackage{url}            
\usepackage{booktabs}       
\usepackage{amsfonts}       
\usepackage{nicefrac}       
\usepackage{microtype}      
\usepackage{xcolor}         

\usepackage{multirow}
\usepackage{siunitx}
\usepackage{graphicx}
 
\title{\textit{Compliance-to-Code}: Enhancing Financial Compliance Checking via Code Generation}

\author{Siyuan Li\textsuperscript{1,*} \quad Jian Chen\textsuperscript{1,* \textdagger}\quad{\bf Rui Yao\textsuperscript{1}} \quad {\bf Xuming Hu\textsuperscript{1}}\quad {\bf Peilin Zhou\textsuperscript{1}} \quad{\bf Weihua Qiu\textsuperscript{2}} \\  \quad{\bf Simin Zhang\textsuperscript{1}} \quad{\bf Chucheng Dong\textsuperscript{3}} \quad{\bf Zhiyao Li\textsuperscript{1}} \quad{\bf Qipeng Xie\textsuperscript{1}} \quad {\bf Zixuan Yuan}\textsuperscript{1,\textdagger}\\
        \textsuperscript{1}Hong Kong University of Science and Technology (Guangzhou) \\
        \textsuperscript{2}Sun Yat-Sen University
        \textsuperscript{3}University of California, Riverside \\
    \{sli974, jchen524, ryao663, pzhou460, szhang420, zli632\}@connect.hkust-gz.edu.cn, \\
        qiuwh9@mail2.sysu.edu.cn, cdong040@ucr.edu, 
        qxieaf@connect.ust.hk \\ 
        \{xuminghu, zixuanyuan\}@hkust-gz.edu.cn 
        \\ {\footnotesize $^{*}$Equal contribution.\; $^{\dagger}$Corresponding authors.}
}

\begin{document}

\maketitle

\begin{abstract}
  Nowadays, regulatory compliance has become a cornerstone of corporate governance, ensuring adherence to systematic legal frameworks. 
At its core, financial regulations often comprise highly intricate provisions, layered logical structures, and numerous exceptions, which inevitably result in  labor-intensive or comprehension challenges. 
To mitigate this, recent Regulatory Technology (RegTech) and Large Language Models (LLMs) have gained significant attention in automating the conversion of regulatory text into executable compliance logic.
However, their performance remains suboptimal particularly when applied to Chinese-language financial regulations, due to three key limitations: (1) incomplete domain-specific knowledge representation, (2) insufficient hierarchical reasoning capabilities, and (3) failure to maintain temporal and logical coherence.
One promising solution is to develop a domain specific and code-oriented dataset for model training. 
Existing datasets such as LexGLUE, LegalBench, and CODE-ACCORD are often English-focused, domain-mismatched, or lack fine-grained granularity for compliance code generation. To fill these gaps, we present \textit{Compliance-to-Code} \footnote{\url{https://github.com/AlexJJJChen/Compliance-to-Code}}, the first large-scale Chinese dataset dedicated to financial regulatory compliance. Covering 1,159 annotated clauses from 361 regulations across ten categories, each clause is modularly structured with four logical elements-subject, condition, constraint, and contextual information-along with regulation relations. We provide deterministic Python code mappings, detailed code reasoning, and code explanations to facilitate automated auditing. To demonstrate utility, we present \textit{FinCheck}: a pipeline for regulation structuring, code generation, and report generation. \textit{Compliance-to-Code} establishes a new benchmark for LLM-based compliance automation. Experimental evaluation shows that GLM-4-9B-0414 achieves the best performance on key task in regulation structuring and DeepSeek-R1-0528 performs the best in compliance code generation tasks.
\end{abstract}

\input{chapter/1-Introduction}
\input{chapter/2-RelatedWorks}
\input{chapter/3-Compliance-to-Code-Dataset}
\input{chapter/4-FinCheck-Pipeline}

\input{chapter/5-Experimental-Setup}
\input{chapter/6-Results-and-Analysis}
\input{chapter/7-Conclusion}

\bibliographystyle{ACM-Reference-Format} 
\bibliography{bib/agent,bib/general_references,bib/KG,bib/legal_domain} 

\appendix

\section{Acknowledgements}
We would like to express our deepest gratitude to Prof. Zixuan Yuan, our corresponding author, for his exceptional wisdom, visionary guidance, and steadfast mentorship throughout the project. We are grateful to Dr. Yutao Huang (Peking University) for policy consulting, and to Yuting Qiu for legal advice. 
We thank Siyuan Li for leading the dataset construction and experiments, ensuring data quality and experimental accuracy. Jian Chen served as the project lead and contributed to the overall research framework, dataset and experimental design, manuscript preparation, and initial project coordination. Rui Yao contributed to both dataset construction and experimental implementation. Weihua Qiu, Simin Zhang, Chucheng Dong, and Zhiyao Li assisted in dataset creation. Xuming Hu, Peilin Zhou and Qipeng Xie contributed to the revision of the manuscript. We thank Qipeng Xie for providing computational resources. 

This work is supported by the Guangzhou-HKUST(GZ) Joint Funding Program (No. 2024A03J0630) and NSFC (No. 62402413).

\input{chapter/8-License-Accessibility-DataUsage-EthicsStatements}
\input{chapter/9-AnnotationSchema}
\input{chapter/10-ErrorAnalysis}
\input{chapter/11-Prompt-of-Instruction-tuning}

\end{document}

%% file: chapter/1-Introduction.tex
\section{Introduction}

\begin{figure*}
    \centering
    \includegraphics[width=\textwidth]{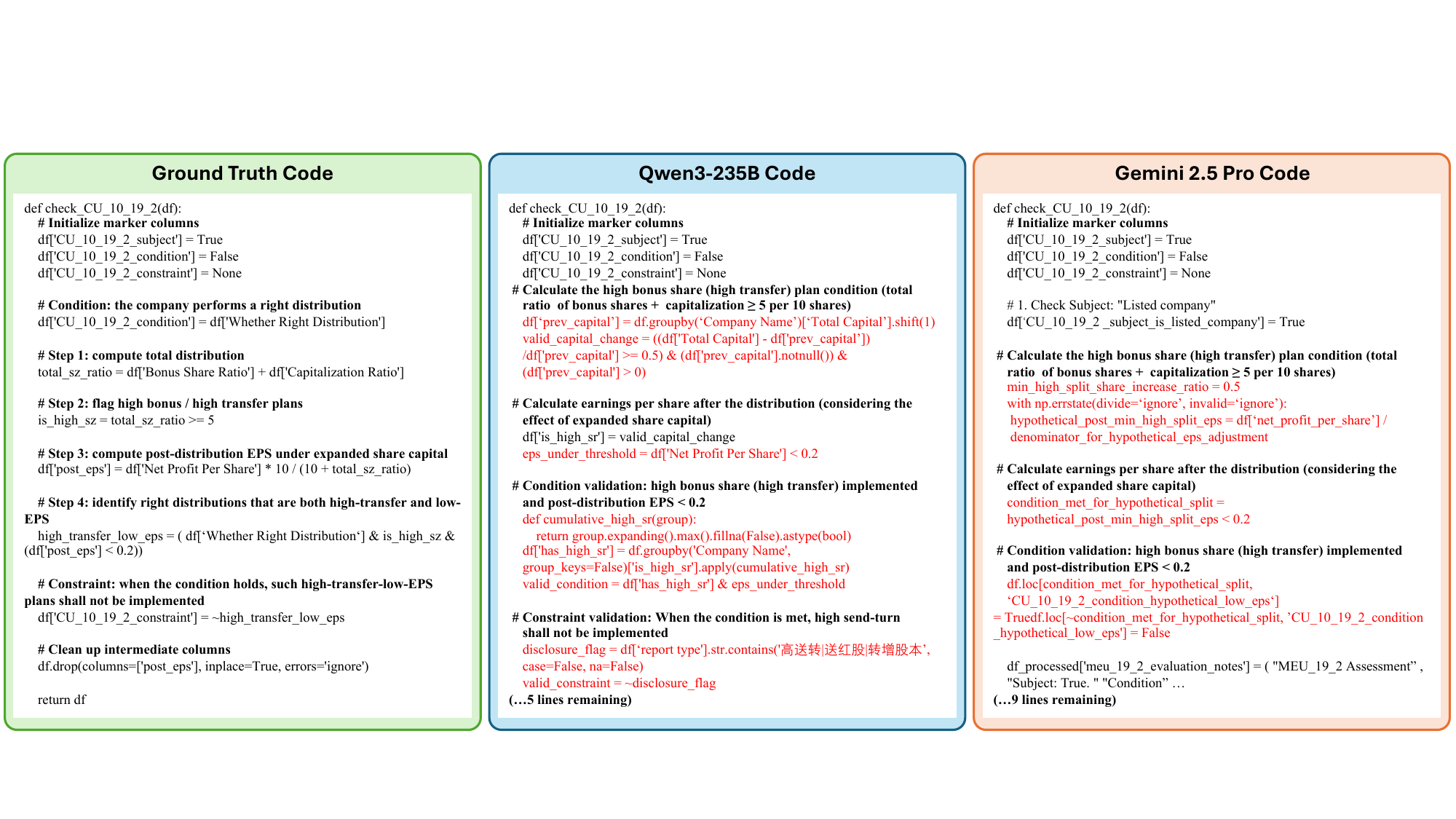}
    \caption{Illustration of error in LLM-generated compliance logic. Code in red color means error.}
    \label{fig:preliminary}
\end{figure*}

Corporate compliance with financial regulations is regarded as a critical mechanism through which institutions ensure their operations adhere to complex legal frameworks and industry standards \cite{depamphilis2019mergers}. 
Modern financial regulations are notoriously elaborate, featuring multilayered conditions, exceptions, quantitative thresholds, and numerous cross-references, all of which require accurate interpretation and operationalization \cite{butler2019understanding}. 
Traditionally, compliance checking relies heavily on legal, risk, and audit teams, incurring high costs and lengthy processing times. 
As the volume and complexity of regulations continue to grow, these manual approaches have become increasingly unsustainable~\cite{weirich2025accounting, martin2024better}.

Regulatory Technology (RegTech) aims to automate compliance management by leveraging advanced computational techniques~\cite{charoenwong2024regtech}. With the emergence of large language models (LLMs), substantial progress has been made in various legal text processing tasks, such as information retrieval, classification, and summarization~\cite{guha_legalbench_2023, kant2025towards}. However, directly employing LLMs for compliance checking poses significant challenges. Due to the inherently black-box nature of LLMs and the non-determinism of their outputs, the generated answers may lack consistency and transparency. By contrast, executable code offers clear, verifiable compliance decisions. This raises a fundamental question: \textit{Can LLMs bridge the gap between natural language regulations and executable compliance code, thereby enabling fully automated compliance auditing?} To explore this, we conducted preliminary experiments with state-of-the-art LLMs, including Gemini-2.5-Pro~\cite{gemini2025} and Qwen3-235B~\cite{qwen3_2025}. As illustrated in Figure~\ref{fig:preliminary}, our results indicate that existing LLMs still exhibit significant limitations in regulation-to-code tasks:

\begin{itemize}
    \item \textbf{Incomplete Domain Knowledge:} LLMs may conflate domain concepts or overlook essential calculation steps. For instance, Gemini-2.5-Pro detected stock distributions by monitoring overall capital changes, ignoring whether such changes resulted strictly from bonus shares, and miscalculated diluted EPS by skipping post-distribution adjustments, resulting in flawed compliance judgments.

    \item \textbf{Shallow Structural Reasoning:} LLMs often fail to capture the precise data granularity or semantics required for financial compliance logic. For example, Qwen3-235B designed rule checks around hypothetical or missing input fields instead of leveraging explicit, available columns (e.g., "bonus share ratio"), leading to incomplete or irrelevant condition detection.
    \item \textbf{Temporal and Logical Inconsistencies:} Both models demonstrate limitations in tracking event causality and enforcing regulatory constraints over time. Qwen3-235B defaulted to over-simplified constraint satisfactions in the absence of explicit disclosure data, while Gemini-2.5-Pro wrongly applied event flags across all future periods and relied on ambiguous keyword searches in announcements, failing to uphold event-specific, rule-based compliance.
\end{itemize}

We attribute these weaknesses primarily to the insufficient domain-specific knowledge of LLMs, and inability to modularize regulatory content, and the absence of appropriate code-oriented training data. Previous research has attempted to address these issues by constructing datasets for improved LLM evaluation in compliance code generation. Existing benchmarks-such as LexGLUE \cite{chalkidis2022lexglue}, LegalBench \cite{guha_legalbench_2023}, and MAUD \cite{wang2023maud}-focus on tasks like legal text classification, clause extraction, or high-level reasoning. However, they do not modularize financial regulations, and therefore are inadequate for the end-to-end compliance code generation scenario~\cite{kuccuk2025computational}. Other datasets, such as CODE-ACCORD \cite{hettiarachchi2025code}, attempt to formalize condition-action relationships within regulations, but are largely domain-agnostic and not tailored to financial compliance, which frequently relies on specialized terminology. Furthermore, CODE-ACCORD \cite{hettiarachchi2025code} and similar datasets do not effectively reduce model hallucinations in this domain and are predominantly in English, limiting their applicability to diverse regulatory environments. Consequently, there is a lack of a high-quality, Chinese-language dataset for financial compliance checking-a critical resource for advancing LLMs' understanding, reasoning, and automation capabilities in regulatory contexts.

To address the lack of automated tools for Chinese financial regulatory compliance, we present the \textit{Compliance-to-Code} dataset-a large-scale, richly annotated resource tailored for structured understanding and code-based reasoning in this domain.

\textbf{Dataset Construction and Features:}
The \textit{Compliance-to-Code} dataset is built from a systematic analysis of 361 authoritative Chinese financial regulations, spanning ten regulatory categories. We extract and decompose 1,159 representative legal clauses into four fundamental logical components-\texttt{subject}, \texttt{condition}, \texttt{constraint}, and \texttt{contextual information}-to create fine-grained Compliance Units (CUs). We also annotate relationships among CUs to support advanced compositional and cross-reference reasoning. Crucially, we selected CUs amenable to computational assessment, annotated them with executable Python modules, and used these as templates to create 307 coding tasks. Each code example includes detailed Chain-of-Thought reasoning and explanatory comments, enhancing transparency and reproducibility.

\textbf{\textit{FinCheck} Pipeline: Automated Compliance Checking.} 
To demonstrate practical utility, we develop \textit{FinCheck}, a modular end-to-end pipeline leveraging the Compliance-to-Code dataset (see Fig.~\ref{fig:example}). \textit{FinCheck} includes: (1) a structure predictor that parses free-text regulations into CUs; (2) a code generator translating CU structures into executable Python code; (3) an information retriever fetching relevant company data; and (4) a report generator producing user-friendly compliance assessments. This pipeline provides traceable, automated compliance checking.

\textbf{Experimental Results:} 
We benchmark recent LLMs for regulation structuring and code generation tasks using our dataset's detailed annotations. DeepSeek-R1-0528 \cite{guo2025deepseek} achieves the best performance on code generation. With Supervised Fine-Tuning (SFT), GLM-4-9B-0414 notably improves its regulation structuring capabilities.

Summary of Contributions:

1. \textit{Compliance-to-Code} dataset: We developed the first large-scale Chinese financial regulation dataset structured for code-based compliance reasoning, with 1,159 annotated clauses, 307 Python modules, and reasoning steps.

2. \textit{FinCheck} pipeline: \textit{FinCheck} turns the dataset's structure into a practical, auditable automation solution, supporting robust benchmarking.

3. Comprehensive evaluation: Our experiments provide strong baselines and highlight both challenges and opportunities for code-driven financial compliance in China.

%% file: chapter/2-RelatedWorks.tex
\section{Related Works}
\subsection{RegTech Methods for Financial Compliance}

Regulatory Technology (RegTech) leverages computational tools to automate financial compliance~\cite{butler2019understanding}. Early RegTech solutions relied on deterministic, rule-based systems and expert models (e.g., GEM~\cite{hassan2009governance}) using formal logic and standards like XBRL~\cite{jeyasingh2023impact}. While precise and explainable, these symbolic systems struggled with scalability and adaptability amid evolving regulations.

Recent advances in AI, notably Large Language Models (LLMs), have enabled data-driven approaches for parsing regulations and automating compliance workflows~\cite{sun2025compliance, jiang2025intellichain}. These neural models offer broader generalization but face challenges with domain specificity, rapid regulatory change, and the complexity of legal logic. Key limitations remain in transparency, interpretability, and auditability-essential for compliance contexts~\cite{wang2024report}. Even Retrieval-Augmented Generation (RAG) \cite{wiratunga2024cbr}, designed explicitly to adapt to frequent regulatory updates, continues to encounter substantial difficulties due to suboptimal retrieval accuracy and the generator's limited capacity for complex logical reasoning in compliance checking.

To address this, hybrid pipelines combining neural models with symbolic reasoning or human oversight have become prominent~\cite{wang2024report, sun-etal-2025-compliance}. Our work follows this trajectory, presenting a code-centric approach for LLM-driven compliance automation that enhances rigor, reliability, and legal traceability beyond brittle rule-based pipelines.

\subsection{Related Datasets and Benchmarks}

Despite active research in RegTech, there is a critical lack of datasets mapping real regulatory text to executable compliance logic in the financial sector. Most existing resources address subtasks such as document classification (e.g., LEDGAR~\cite{tuggener-etal-2020-ledgar}, LexGLUE~\cite{chalkidis2022lexglue}), judgment prediction, or QA (e.g., LegalBench~\cite{guha_legalbench_2023}, BillSum~\cite{kornilova-eidelman-2019-billsum}), but lack annotations for code-level reasoning or logic synthesis.

Datasets closer to financial regulation, like ObliQA~\cite{gokhan2024regnlp} and BSARD~\cite{lotfi2025bilingual}, only support question answering or retrieval. CODE-ACCORD~\cite{hettiarachchi2025code} stands out as a logic-annotated resource, but it is limited to building regulations and has narrow domain coverage. Crucially, no existing benchmark, especially in Chinese language systematically couples financial statutes with executable, expert-annotated compliance logic, impeding progress towards true compliance to code. Our work directly addresses this foundational gap.

%% file: chapter/3-Compliance-to-Code-Dataset.tex
\section{\textit{Compliance-to-Code} Dataset}



\subsection{Data Collection and Pre-processing}

\textit{Compliance-to-Code} Dataset is systematically curated from official regulatory documents published by the Beijing Stock Exchange (BSE)\footnote{\url{https://www.bse.cn/rule/secnotice_list.html}}, with a focus on compliance obligations, internal controls, and risk points for publicly listed companies in China. In comparison to the regulations issued by the China Securities Regulatory Commission (CSRC)\footnote{\url{http://www.csrc.gov.cn}}, the BSE provides more detailed and supplementary rules, increasing both the granularity and complexity of regulatory interpretation. We specifically gathered authoritative documents classified as Departmental Rules and Normative Documents, which constitute the regulatory backbone of corporate compliance obligations and reporting requirements. The regulatory corpus reflects all in-force documents as of January 1, 2025, providing a stable but comprehensive snapshot for benchmarking compliance automation.

We identified ten core thematic areas central to corporate compliance program design (see Table~\ref{tab:regulatory_aspects}). Thematic selection was motivated by three criteria: (i) relevance to regulatory risk exposures, (ii) frequency and significance of compliance controls, and (iii) interpretative complexity, thus aligning with both enterprise compliance management frameworks and prevailing regulatory guidance. In addition, considering the dynamic nature of regulatory changes, we prioritized rules with broad applicability and removed highly exceptional or idiosyncratic provisions. For extremely ambiguous regulations-where even expert legal opinion is unavailable-these were deliberately excluded to enhance dataset reliability. 

The initial preprocessing stage involved robust legal text extraction, normalization, and standardization, ensuring accurate and complete conversion from diverse official document formats into a unified legal text corpus.

\subsection{Data Annotation}
Drawing inspiration from RegGPT \cite{wang2024reggpt}, we observe that regulations are typically composed of six core elements: meta data (entity/property), constraint, condition, measure, scope, and external references. To facilitate more effective translation of regulatory rules into executable code by LLMs, we propose several refinements to this schema. Specifically, we merge the elements of measure and constraint under the unified category of \texttt{Constraint}, relabel external references as \texttt{Contextual Information}, and rename meta data to  \texttt{Subject}. 

Based on these modifications, we introduce the concept of a \textit{Compliance Unit (CU)}-the minimal actionable regulatory statement, which can be a rule, requirement, or prohibition. Each CU is structured by four key components:  \texttt{Subject}, \texttt{Condition}, \texttt{Constraint}, and \texttt{Contextual Information} (e.g., source, applicability, or exceptions).

To capture compliance logic, we annotate four directed \textit{Inter-Unit Relations}: \texttt{refer to} (CU references another), \texttt{exclude} (CU negates another), \texttt{only include} (logic applies only if another CU is met), and \texttt{should include} (compliance depends on satisfying another CU). See Appendix \ref{app:cu_annotation} for definitions and examples.

Annotation leveraged a multidisciplinary team in securities law, financial compliance, and audit automation, operating in four stages:

\begin{enumerate}
    \item \textbf{CU Identification:} Legal/compliance experts segmented regulatory text into CUs, tagging each field and clarifying compliance intent, resolving ambiguities with compliance officers and legal counsel.
    \item \textbf{Inter-Unit Relation Annotation:} Experts identified logical relations between CUs.
    \item \textbf{Compliance Logic Archetypes:} Domain specialists abstracted 68 distinct logic patterns from annotated CUs and formulated Python-based automation tasks mirroring realistic compliance scenarios. To model regulatory volatility, we varied numerical values and constraints in these code templates, simulating clause amendments and threshold changes.
    \item \textbf{Reasoning and Validation:} Each task includes a detailed reasoning trace mapping regulation to code, with dual expert review to ensure accuracy.
\end{enumerate}

We reformulated each programming problem as a code completion task and categorized them into three difficulty levels: simple, medium, and difficult. 
For \textbf{Simple} problems, the LLM needed to fill in one or two missing lines of code; for \textbf{Medium} problems, three lines; and for \textbf{Difficult} problems, the LLM was tasked with generating the entire solution from scratch.
DeepSeek-R1 \cite{guo2025deepseek} provided candidate extractions and code suggestions, but all outputs underwent independent expert review and adjudication. 

Rigorous multi-layered quality assurance encompassed: $\;$ (i) Adherence to expert-developed annotation guidelines, (ii) Inter-annotator agreement ($\kappa{=}0.866$ on 15\% subset, 180 CUs), (iii) Automated artifact checks and extensive unit testing for code, and (iv) Expert audit of 68 randomly selected automation tasks against source obligations and industry best practice. More details of data annotation are shown in the supplementary material.

\subsection{Dataset Statistics}
\begin{figure}
    \centering
    \includegraphics[width=1\linewidth]{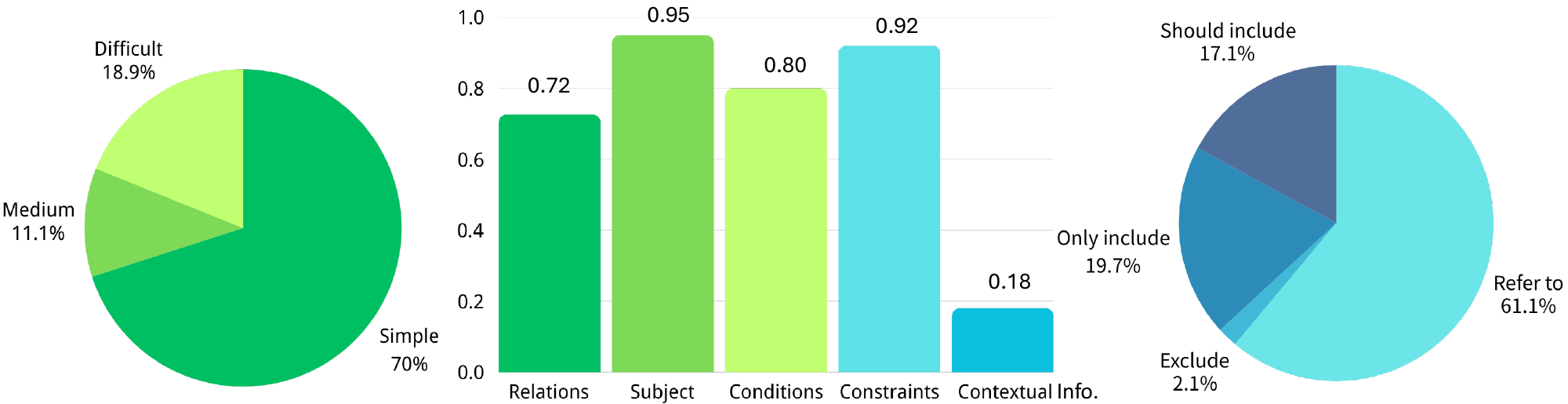}
    \caption{Distribution of Dataset Difficulty Levels, Average Number of Compliance Unit in One Clause and Inter-Unit Relation Types.}
    \label{fig:distribution}
\end{figure}

The \textit{Compliance-to-Code} dataset consists of: (1) clean regulatory texts; (2) structured CUs and their inter-unit relations; (3) Python-coded compliance automation tasks; (4) expert-validated COT code generation analyses. The base covers 361 regulations (10 domains) with 53,340 words.

Final annotation produced 1,159 CUs (from 1,226 candidates), and 864 directed inter-unit relations - \texttt{refer to} (61.1\%), \texttt{exclude} (2.1\%), \texttt{only include} (19.7\%), \texttt{should include} (17.1\%). Each CU has, on average, 0.72 relations, 0.95 subjects, 0.80 conditions, and 0.92 constraints and 0.18 contextual information. (in Figure \ref{fig:distribution})

These were mapped to 307 distinct executable compliance automation tasks (Python), each code module has a COT reasoning step (average 8 steps/2145 tokens). 

%% file: chapter/4-FinCheck-Pipeline.tex
\section{\textit{FinCheck}: A Structured Compliance Checking Pipeline}

\begin{figure*}[t]
\centering
\includegraphics[width=1\linewidth]{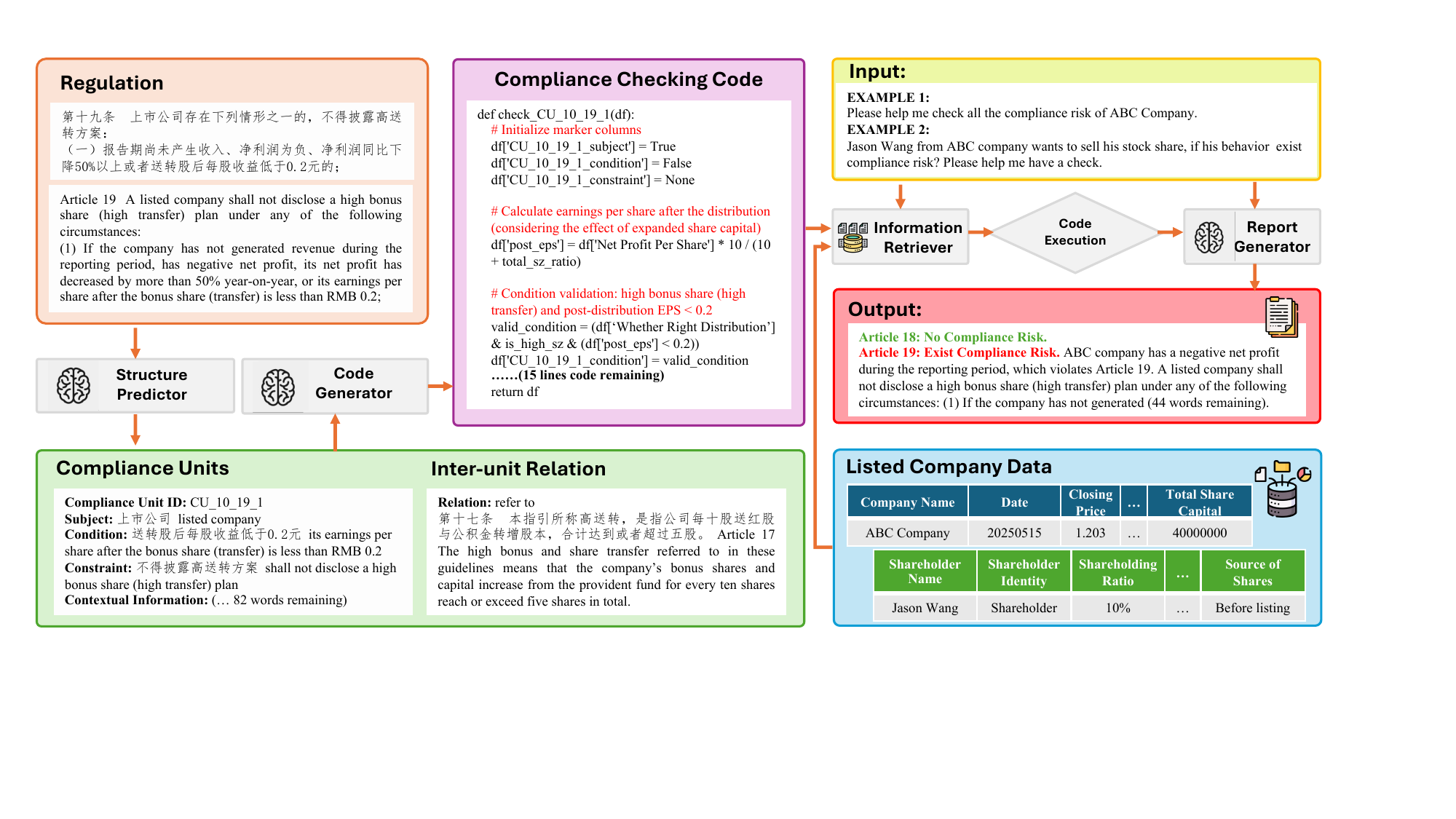}
\caption{\textit{FinCheck} Compliance Checking Pipeline. The framework processes natural language regulations by first using a Structure Predictor to extract key compliance units. These units are then fed into a Code Generator to create verification code. For a specific case, user input triggers an Information Retriever to fetch relevant company data. The generated code is executed with this data, and a Report Generator summarizes the outcome before the final verification result is shown to the user.}
\label{fig:example}
\end{figure*}

\textit{FinCheck} is a systematic pipeline for automating compliance checks against financial regulations written in natural language. Its core idea is to convert regulatory text into machine-executable code, allowing transparent and reproducible compliance checking on real-world financial data. As shown in Figure~\ref{fig:example}, \textit{FinCheck} comprises four modular components:

\textbf{Structure Predictor:} Given a regulatory document in natural language, a fine-tuned large language model (LLM) is employed to extract \emph{Compliance Units} (CUs) and relations required for compliance assessment. These CUs, along with logical relations among them, are organized into a structured, machine-readable format that makes the regulatory logic explicit.

 \textbf{Code Generator:} The structured representation of the regulation (i.e., CUs and their relations) serves as input to a \textbf{\textit{Structured Synthesis LLM (SS-LLM)}}, also fine-tuned for this task, which then generates Python code implementing the compliance checking logic. To enhance reliability and reasoning ability, the LLM is fine-tuned via supervised learning with step-by-step reasoning.
 The resulting code realizes the regulatory requirements as formal programmatic checks.

 \textbf{Information Retriever:} For any given user query, this component retrieves relevant company or client data (based on Wind Information\footnote{\url{https://www.wind.com.cn/mobile/EDB/en.html}}) from structured sources, preparing it as input for the generated compliance-checking code. We utilize Python-based dataframe operations for efficient and flexible data retrieval aligned with the code's requirements.

\textbf{Report Generator:} The outcome of code execution is summarized using a dedicated LLM, which is prompted to ensure clear, interpretable, and standardized reporting. This report provides end-users with both the compliance result and a transparent trace of the logical reasoning behind the decision.

%% file: chapter/5-Experimental-Setup.tex
\section{Experimental Setup}
\label{sec:experiments}

\subsection{Core Tasks}
\label{subsec:tasks_and_objectives}
We investigate three key tasks reflecting the main stages of the compliance checking pipeline:
\begin{itemize}
    \item \textbf{Task 1: Regulation-to-Structure (R2S)}: Automatically parses raw financial regulatory text into structured Compliance Units (CUs) and their logical relations.
    \item \textbf{Task 2: Structure-to-Code (S2C)}: Converts (gold or predicted) CUs and relations into functionally correct, auditable Python code implementing compliance logic.
    \item \textbf{Task 3: Regulation-to-Code (R2C)}: Directly generates executable compliance code from regulatory text, representing the end-to-end challenge.
\end{itemize}

\subsection{Baselines}
Since there are currently no models specifically designed to convert regulations into code, we are unable to employ any models directly aimed at Task 2. Furthermore, although RegGPT \cite{wang2024reggpt} demonstrates promising performance for Task 1, it is not open-sourced, which limits our ability to use it in our experiments. As a result, we focus our experiments on general LLMs.

\textbf{Baseline 1: Direct Prompting LLM (DP-LLM).} Instruction-tuned LLMs applied to Task 2 and 3; evaluated with zero-shot and one-shot prompts. The prompt is shown in Appendix \ref{app:prompt}.

\textbf{Baseline 2: Chain-of-Thought LLM (CoT-LLM).} As above, but prompts elicit step-by-step reasoning prior to code generation, using demonstration of chain-of-thought examples.

\subsection{Dataset Splits}

For our first task, regulation-to-structure, the dataset comprises 1,159 compliance units (CUs), which are divided into training (70\%), validation (15\%), and test (15\%) sets with stratification.

For our second task, \textit{structure-to-code}, we implemented a strict domain generalization benchmark. We partitioned the available compliance control domains into two disjoint sets. The training set is composed of 249 'Difficult' tasks (81.1\%) sourced from the first set of domains. The remaining 58 tasks (18.9\%) were drawn from the second. To enable a fine-grained evaluation of model performance, we systematically expanded these 58 core tasks into a hierarchical, 307-item test set. This suite, composed of the original 58 'Difficult' tasks, plus 34 'Medium' and 215 'Simple' derived variants, allows for a detailed analysis of how model capabilities vary with complexity. 


\subsection{Training Protocols}

\textbf{Base Models}: For Task 1, we utilize Qwen3-8B \cite{qwen3_2025}, DeepSeek-R1 and DeepSeek-R1-Distill-Qwen-7B \cite{guo2025deepseek}, GLM-4-9B-0414 \cite{thudm_glm4_0414_2025}, GPT-4.1 \cite{openai_gpt4_1_2025}, Gemini-2.5-pro \cite{gemini2025} and Claude-3.7-Sonnet \cite{anthropic_claude3_7_2025}. We also included Fin-R1 \cite{liu2025finr1largelanguagemodel}, a 7B financial domain model created by fine-tuning Qwen2.5-7B-Instruct on financial reasoning tasks. For Tasks 2 and 3, which focus on code generation, we additionally include Qwen2.5-Coder-7B-Instruct \cite{hui2024qwen2} and DeepSeek-Coder-6.7B-Instruct \cite{guo2024deepseek} in our evaluation.

\textbf{Supervised Fine-Tuning (SFT)}: We train using causal language modeling (next-token prediction), AdamW optimizer, cosine learning rate schedule, batch size 1, weight decay 0.01. The learning rate is set at $1 \times 10^{-4}$, epochs $1$, top\_p $0.75$, and temperature $0$. For LoRA, we use rank 8, $\alpha=32$, and dropout $0.05$. We employ \texttt{Swift} \footnote{\url{https://github.com/modelscope/ms-swift}} for efficiency. The prompt is shown in Appendix \ref{app:prompt}.


All training and inference are conducted on 16x NVIDIA A800 80GB GPUs for 14 days.

\subsection{Evaluation Metrics}
For Task 1 (Regulation-to-Structure), evaluation includes \textit{CU Field Extraction}, assessed using entity-level Precision, Recall, and F1-score for Subject, Condition, and Constraint spans. The temperature was set to 0 for decoding. \textit{CU Boundary Identification} is evaluated with Precision, Recall, and F1-score for correctly identifying the start and end tokens of each CU. \textit{Relation Extraction} performance is measured by Precision, Recall, and F1-score for identifying relation types and their CU arguments.

For Tasks 2 \& 3 (Code Generation), we primarily evaluate \textit{Functional Correctness}, measured by Pass@1 and Pass@5. Specifically, Pass@k represents the percentage of generated code samples that successfully compile and pass all associated unit tests within the first k attempts. The temperature was set to 0.6, except for DeepSeek-R1-0528, where it was set to 0.0 due to cost considerations. As a secondary metric, we report Code Similarity, quantified by CodeBLEU~\cite{ren2020codebleu}, which comprehensively assesses the lexical, syntactic, and semantic similarity between generated code and the reference solutions.


Qualitative and Auditability-focused metrics, primarily used for SS-LLM, include \textit{Commentary Traceability}, evaluated as the percentage of input CUs and relations explicitly referenced in the generated code comments (manual evaluation on a subset). \textit{Explanation Quality} is assessed via domain expert ratings (1-5 Likert scale) for clarity, faithfulness, and completeness of generated CoT explanations (for CoT-LLM) and code commentary (for SS-LLM). Finally, \textit{Error Analysis} involves manual classification of failure modes for incorrect generations on the test set.

%% file: chapter/6-Results-and-Analysis.tex
\section{Results and Analysis}
\label{sec:results}

\subsection{Task 1 Performance: Advancing Automated Structuring of Complex Regulatory Text}
The efficacy of our end-to-end pipeline hinges on accurately transforming convoluted regulatory text into structured, actionable CUs and their inter-unit relations. Table~\ref{tab:task1_detailed_restructured} presents a comprehensive evaluation of models fine-tuned on \textit{Compliance-to-Code} against robust few-shot baselines for Task 1. Our central hypothesis posits that domain-specific fine-tuning, particularly with models adept at capturing nuanced semantics in Chinese legal text is pivotal for this task.

\begin{table}[t]
\centering
\caption{Task 1 Performance (Precision, Recall, F1 \%) on the Test Set. CU Fields denotes average scores for Subject, Condition, Constraint. CU Identification concerns overall span detection. Best model scores are in \textbf{bold}. All Fine-tuned Models are performed using Supervised Fine-Tuning (SFT).}
\label{tab:task1_detailed_restructured}
\renewcommand{\arraystretch}{1.2}
\resizebox{0.6\textwidth}{!}{%
\begin{tabular}{@{}l S S S S S S S S S @{}}
\toprule
\multicolumn{1}{c}{\multirow{3}{*}{Model}} &
\multicolumn{3}{c}{CU Field Extraction} & \multicolumn{3}{c}{CU Boundary Identification} \\
\cmidrule(lr){2-4} \cmidrule(lr){5-7} \cmidrule(lr){8-10}
& {P (\%)} & {R (\%)} & {F1 (\%)} & {P (\%)} & {R (\%)} & {F1 (\%)} \\
\midrule
\multicolumn{10}{@{}l}{\textit{Few-shot Performance of Base Models (Task 1: Regulation to Structure, one-shot)}} \\
Qwen3-8B & 60.74 & 56.88 & 58.74 & 28.57 & 28.57 & 28.57 \\
GLM-4-9B-0414 & 48.01 & 44.96 & 46.44 & 23.21 & 23.21 & 23.21 \\
GPT-4.1 & 64.17 & 63.20 & 63.69 & 40.18 & 40.18 & 40.18 \\
DeepSeek-V3-0324 & 71.05 & 68.01 & 69.50 & 49.11 & 49.11 & 49.11 \\
Claude-3.7-Sonnet & 65.28 & 64.49 & 64.89 & 41.07 & 41.07 & 41.07 \\
Gemini-2.5-pro & 71.33 & 70.59 & 70.96 & 48.21 & 48.21 & 48.21 \\
\textbf{DeepSeek-R1-0528} & \textbf{72.01} & \textbf{70.97} & \textbf{71.49} & \textbf{50.00} & \textbf{50.00} & \textbf{50.00} \\
Fin-R1 & 58.08 & 53.92 & 55.92 & 13.39 & 13.39 & 13.39 \\

\midrule

\multicolumn{10}{@{}l}{\textit{Fine-tuned Models (Task 1: Regulation to Structure)}} \\
\textbf{GLM-4-9B-0414} & \textbf{72.10} & \textbf{71.27} & \textbf{71.68} & \textbf{37.50} & \textbf{37.50} & \textbf{37.50} \\
Qwen3-8B & 63.45 & 63.03 & 63.24 & 31.25 & 31.25 & 31.25 \\
DeepSeek-R1-Distill-Qwen-7B & 54.97 & 53.28 & 54.11 & 23.21 & 23.21 & 23.21 \\
\bottomrule
\end{tabular}
} 
\end{table}

Our results indicate that domain-specific supervised fine-tuning (SFT) is an effective strategy. This is demonstrated by the \textbf{GLM-4-9B-0414} model. As a 9B parameter model, its performance in CU field extraction after SFT (\textbf{71.68\% F1}) surpasses that of powerful proprietary models in a few-shot setting, including \textbf{DeepSeek-R1-0528 (71.49\% F1)} and Gemini-2.5-pro (70.96\% F1). This outcome suggests that a smaller, well-tuned model can outperform larger, general-purpose counterparts on specialized tasks, representing an efficient path to achieving a high level of performance.


\begin{table*}[t]
\centering
\caption{Performance of Various Foundation Models before and after SFT for Task 2 (Structure-to-Code, Gold Struct). All models are fine-tuned using our SS-LLM-SFT approach on the \textit{Compliance-to-Code} dataset. Best results in each category are highlighted in \textbf{bold}.}
\label{tab:model_performance_comparison}
\sisetup{round-mode=places, round-precision=2, table-format=2.2}
\resizebox{1.0\textwidth}{!}{%
\begin{tabular}{@{}l S[table-format=2.2] S[table-format=2.2] S[table-format=2.2] S[table-format=2.2] S[table-format=2.2] S[table-format=2.2] S[table-format=2.2] S[table-format=2.2] S[table-format=2.2] S[table-format=2.2] S[table-format=2.2] S[table-format=2.2]@{}}
\toprule
\multicolumn{1}{c}{\multirow{2}{*}{Model}} & \multicolumn{3}{c}{Overall (N=307)} & \multicolumn{3}{c}{Simple (N=215)} & \multicolumn{3}{c}{Medium (N=34)} & \multicolumn{3}{c}{Difficult (N=58)} \\
\cmidrule(lr){2-4} \cmidrule(lr){5-7} \cmidrule(lr){8-10} \cmidrule(lr){11-13}
& {{Pass@1}} & {{Pass@5}} & {CodeBLEU} & {{Pass@1}} & {{Pass@5}} & {CodeBLEU} & {{Pass@1}} & {{Pass@5}} & {CodeBLEU} & {{Pass@1}} & {{Pass@5}} & {CodeBLEU} \\
\midrule
\multicolumn{13}{l}{\textit{Fine-tuned Model}} \\
DeepSeek-R1-Distill-Qwen-7B (SFT) & 27.52 & 56.09 & 61.66 & 36.00 & 68.81 & 67.68 & 9.12 & 27.77 & 46.96 & 6.36 & 24.80 & 47.66 \\
GLM-4-9B-0414 (SFT) & 39.72 & 65.94 & 70.68 & 51.11 & 77.12 & 78.60 & 14.71 & 45.12 & 57.70 & 11.45 & 35.92 & 48.34 \\
Qwen2.5-Coder-7B-Instruct (SFT) & 35.80 & 60.15 & \textbf{71.57} & 48.06 & 76.02 & \textbf{79.51} & 13.24 & 40.17 & \textbf{60.27} & 2.73 & 11.62 & 48.09 \\
\textbf{Qwen3-8B (SFT)} & \textbf{54.58} & \textbf{80.19} & 69.30 & \textbf{65.05} & \textbf{88.15} & 76.71 & \textbf{38.24} & \textbf{68.37} & 55.23 & \textbf{24.55} & \textbf{56.96} & \textbf{49.61} \\
deepseek-coder-6.7b-instruct (SFT) & 35.67 & 64.87 & 68.74 & 46.26 & 78.08 & 76.82 & 11.18 & 32.77 & 55.51 & 10.18 & 34.00 & 45.94 \\
\midrule
\multicolumn{13}{l}{\textit{Base Model}} \\
DeepSeek-R1-Distill-Qwen-7B & 18.40 & 48.37 & 57.49 & 23.91 & 60.88 & 63.99 & 7.06 & 22.48 & 41.01 & 4.27 & 16.36 & 42.73 \\
GLM-4-9B-0414 & 41.75 & 67.95 & 70.74 & 52.68 & 78.71 & 78.20 & 22.65 & 54.52 & 59.01 & 11.64 & 35.01 & 49.38 \\
Qwen2.5-Coder-7B-Instruct & 41.67 & 63.43 & 71.96 & 54.83 & 76.99 & 79.91 & 17.06 & 43.78 & 60.01 & 6.36 & 23.55 & 48.88 \\
Qwen3-8B & 60.03 & 82.93 & 69.81 & 68.10 & 88.57 & 77.52 & 51.18 & 75.25 & 56.24 & 34.55 & 66.06 & 48.63 \\
deepseek-coder-6.7b-instruct & 28.47 & 58.29 & 62.75 & 37.63 & 72.24 & 72.62 & 8.53 & 31.87 & 46.80 & 5.64 & 21.13 & 34.74 \\
Fin-R1 & 37.92 & 62.74 & 68.03 & 50.31 & 77.64 & 75.11 & 12.94 & 36.52 & 54.50 & 5.82 & 21.77 & 49.21 \\
\textbf{DeepSeek-R1-0528} & \textbf{73.67} & {-} & \textbf{73.34} & \textbf{81.04} & {-} & \textbf{80.20} & \textbf{70.59} & {-} & \textbf{63.76} & \textbf{47.27} & {-} & \textbf{52.96} \\
\bottomrule
\end{tabular}
} 
\end{table*}

Error analysis pinpoints a primary bottleneck in the propagation of contextual information, particularly within complex legal obligations that span multiple sentences. Models often struggle to maintain and transmit contextual cues across these long-form dependencies, leading to omissions or misinterpretations. Despite these remaining challenges, the structuring phase delivers a reliable foundation for downstream code synthesis, preserving the majority of regulatory semantics and enabling scalable automation for dense financial texts.

\begin{table*}[t]
\centering
\caption{Performance Evaluation of Different Models and Approaches on Code Generation Tasks (Model: Qwen3-8B). Gold Struct implies human-annotated structures. Best results in each main section are in \textbf{bold}.}
\label{tab:qwen3_8b_performance}
\resizebox{1\textwidth}{!}{%
\begin{tabular}{@{}l c c c c c c c c c c c c@{}}
\toprule
\multicolumn{1}{c}{\multirow{2}{*}{Model / Approach}} & \multicolumn{3}{c}{Overall (N=307)} & \multicolumn{3}{c}{Simple (N=215)} & \multicolumn{3}{c}{Medium (N=34)} & \multicolumn{3}{c}{Difficult (N=58)} \\
\cmidrule(lr){2-4} \cmidrule(lr){5-7} \cmidrule(lr){8-10} \cmidrule(lr){11-13}
& {Pass@1} & {Pass@5} & {CodeBLEU} & {Pass@1} & {Pass@5} & {CodeBLEU} & {Pass@1} & {Pass@5} & {CodeBLEU} & {Pass@1} & {Pass@5} & {CodeBLEU} \\
\midrule
\multicolumn{13}{@{}l}{\textit{Prompting Baselines (Task 3: Regulation-to-Code, Model: Qwen3-8B)}} \\
CoT-LLM (one-shot) & 52.29 & \textbf{70.93} & \textbf{69.49} & 66.42 & 87.59 & \textbf{77.53} & \textbf{46.76} & 75.72 & 56.86 & 0.56 & \textbf{2.78} & 46.03 \\
CoT-LLM (0-shot) & 52.69 & 68.34 & 68.68 & 67.27 & 84.44 & 76.89 & 45.88 & \textbf{77.02} & 60.11 & 0.00 & 0.00 & 42.03 \\
DP-LLM (one-shot) & \textbf{52.86} & 70.11 & 69.34 & \textbf{68.03} & 87.00 & 77.32 & 41.47 & 72.54 & 56.64 & \textbf{0.74} & 2.62 & \textbf{46.16} \\
DP-LLM (0-shot) & 51.99 & 70.62 & 68.60 & 66.33 & \textbf{88.16} & 76.93 & 45.59 & 73.87 & \textbf{60.12} & 0.00 & 0.00 & 41.40 \\
\midrule
\multicolumn{13}{@{}l}{\textit{Prompting Baselines (Task 2: Structure-to-Code, Model: Qwen3-8B, Gold Struct)}} \\
CoT-LLM (one-shot) & \textbf{61.00} & \textbf{82.97} & 69.76 & \textbf{70.21} & \textbf{89.15} & 77.22 & \textbf{47.35} & 75.98 & 57.53 & 34.09 & 63.60 & \textbf{48.68} \\
CoT-LLM (0-shot) & 52.08 & 70.53 & 69.02 & 66.71 & 87.75 & 77.40 & 45.59 & \textbf{77.80} & \textbf{60.97} & 0.00 & 0.00 & 41.83 \\
DP-LLM (one-shot) & 60.58 & 82.84 & \textbf{69.83} & 69.36 & 88.98 & \textbf{77.63} & \textbf{47.35} & 73.52 & 56.88 & \textbf{35.09} & \textbf{65.05} & 47.90 \\
DP-LLM (0-shot) & 53.50 & 69.69 & 68.81 & 68.58 & 87.35 & 77.50 & 46.47 & 72.85 & 59.48 & 0.00 & 0.00 & 41.21 \\
\midrule
\multicolumn{13}{@{}l}{\textit{Composed Pipeline (Regulation-to-Structure-to-Code, Model: Qwen3-8B, Predicted Struct)}} \\
CoT-LLM (one-shot) & 54.58 & 69.39 & 69.12 & 70.55 & 87.64 & 77.24 & 43.53 & 66.87 & 55.62 & 0.18 & 0.91 & 46.33 \\
\midrule
\multicolumn{13}{@{}l}{\textit{Reasoning-Augmented Setting (Structure-Reasoning-to-Code, Model: Qwen3-8B, Gold Struct)}} \\
CoT-LLM (one-shot) & 78.10 & \textbf{91.28} & 72.41 & 84.95 & \textbf{94.21} & 79.23 & 62.35 & \textbf{84.34} & 62.09 & \textbf{62.11} & \textbf{84.57} & 53.31 \\
DP-LLM (one-shot) & \textbf{78.13} & 90.28 & \textbf{73.28} & \textbf{85.40} & 94.11 & \textbf{79.96} & \textbf{64.41} & 83.53 & \textbf{63.90} & 59.39 & 80.15 & \textbf{54.17} \\
\bottomrule
\end{tabular}
} 
\end{table*}

\subsection{Tasks 2\&3: Structured Regulations Are Transformative for Compliance Code Generation}

\paragraph{Effect of SFT and Model Choice.}
As shown in Table~\ref{tab:model_performance_comparison}, the effectiveness of supervised fine-tuning (SFT) varies significantly across models. For instance, Qwen3-8B achieves strong performance both as a base model and after SFT, with only marginal differences (e.g., Pass@1: 60.03\% base vs. 54.58\% SFT). In contrast, other models such as DeepSeek-R1-Distill-Qwen-7B and deepseek-coder-6.7b-instruct benefit more noticeably from SFT. This suggests that the pre-training quality and scale of the foundation model play a critical role in code generation task, and that SFT may yield diminishing returns for already highly capable models.

\paragraph{Impact of Structured Inputs and Reasoning.}
Table~\ref{tab:qwen3_8b_performance} demonstrates that structured input and explicit reasoning steps substantially improve code generation performance. Moving from direct prompting on raw regulations (Task 3, Pass@1 $\sim$52.9\%) to providing human-annotated structures (Task 2, Pass@1 $\sim$61.0\%) yields a clear performance boost. Incorporating explicit reasoning (Structure-Reasoning-to-Code) further elevates Pass@1 to 78.1\%. This progression highlights the importance of decomposing complex compliance requirements into structured representations and guiding the model through intermediate reasoning steps. We further investigated in Regulation-to-Structure-to-Code. This pipeline shows an improvement over the Regulation-to-Code baseline, while the performance on "Difficult" tasks does not improve significantly. This underscores a critical challenge: error propagation. Upstream structuring errors propagate to the code generation stage, exposing model limitations in complex, multi-hop legal reasoning.

\paragraph{Prompting Strategies.}
Few-shot prompting and chain-of-thought (CoT) approaches generally outperform zero-shot and direct prompting baselines, especially on more challenging examples. For instance, in the difficult subset, CoT-LLM (one-shot) achieves up to 34.09\% Pass@1 (Task 2), compared to near-zero for zero-shot methods. This underscores the value of providing exemplars and reasoning scaffolds, particularly for complex or ambiguous compliance scenarios.

Overall, our findings indicate that while SFT can be beneficial in code generation for some models, the most significant gains in compliance code generation arise from leveraging structured problem decomposition and explicit reasoning in the prompting process. These strategies are especially crucial for handling complex or difficult cases, and their effectiveness is amplified when combined with strong foundation models.

\subsection{Trade-off Between Compliance Reasoning and Stylistic Alignment}

\paragraph{Epoch-wise Divergence of SFT}
Results in Table~\ref{tab:model_performance_qwen3_epochs_sorted} illustrate a clear divergence between stylistic alignment and compliance reasoning during extended LoRA-based SFT of Qwen3-8B on structure-to-code tasks. Increasing epochs consistently enhance CodeBLEU, reflecting improved mimicry of human coding patterns, but fail to yield corresponding improvements in compliance reasoning as evidenced by declining Pass@1. The simultaneous moderate gain in Pass@5 suggests higher output diversity yet degraded confidence calibration. We conjecture that this phenomenon stems from catastrophic forgetting intertwined with over-adaptation: prolonged training incrementally biases the model toward superficial pattern memorization, consequently compromising deeper logical inference capability. These observations highlight an inherent tension in fine-tuning strategies, suggesting a critical need for regularization techniques or strategic early stopping to preserve compliance reasoning amid stylistic improvements.

\begin{table*}[t]
\centering
\caption{Performance Evaluation of Qwen3-8B across multiple SFT epochs on Code Generation Tasks. 20\% of the original training set was divided into validation set to monitor performance. Best results are highlighted in \textbf{bold}.}
\label{tab:model_performance_qwen3_epochs_sorted}
\sisetup{round-mode=places, round-precision=2, table-format=2.2}
\resizebox{1.0\textwidth}{!}{%
\begin{tabular}{@{}l S[table-format=2.2] S[table-format=2.2] S[table-format=2.2] S[table-format=2.2] S[table-format=2.2] S[table-format=2.2] S[table-format=2.2] S[table-format=2.2] S[table-format=2.2] S[table-format=2.2] S[table-format=2.2] S[table-format=2.2]@{}}
\toprule
\multicolumn{1}{c}{\multirow{2}{*}{Model}} & \multicolumn{3}{c}{Overall (N=307)} & \multicolumn{3}{c}{Simple (N=215)} & \multicolumn{3}{c}{Medium (N=34)} & \multicolumn{3}{c}{Difficult (N=58)} \\
\cmidrule(lr){2-4} \cmidrule(lr){5-7} \cmidrule(lr){8-10} \cmidrule(lr){11-13}
& {\texttt{Pass@1}} & {\texttt{Pass@5}} & {CodeBLEU} & {\texttt{Pass@1}} & {\texttt{Pass@5}} & {CodeBLEU} & {\texttt{Pass@1}} & {\texttt{Pass@5}} & {CodeBLEU} & {\texttt{Pass@1}} & {\texttt{Pass@5}} & {CodeBLEU} \\
\midrule
\multicolumn{13}{l}{\textit{Fine-tuned Model}} \\
Qwen3-8B-1-epoch & \textbf{56.27} & \textbf{80.74} & 69.93 & \textbf{66.18} & \textbf{88.37} & 77.33 & \textbf{41.18} & \textbf{69.14} & 56.98 & \textbf{27.55} & 58.64 & 49.55 \\
Qwen3-8B-3-epoch & 52.18 & 76.39 & 69.99 & 65.05 & 84.99 & 77.90 & 28.53 & 65.56 & 57.23 & 17.45 & 50.10 & 47.54 \\
Qwen3-8B-5-epoch & 49.38 & 75.99 & 70.52 & 62.37 & 85.72 & 78.04 & 19.41 & 54.97 & 57.48 & 18.09 & 51.63 & 49.74 \\
Qwen3-8B-7-epoch & 50.35 & 78.38 & 71.10 & 62.63 & 87.91 & \textbf{78.53} & 21.18 & 54.20 & 58.74 & 21.27 & 56.77 & 50.20 \\
Qwen3-8B-9-epoch & 51.10 & 80.39 & \textbf{71.30} & 63.48 & 87.84 & 78.07 & 24.71 & 64.74 & \textbf{61.11} & 19.91 & \textbf{61.48} & \textbf{51.63} \\
\bottomrule
\end{tabular}
} 
\end{table*}

\subsection{Auditability, Maintainability, and Granular Error Analysis}
Beyond quantitative metrics, the pipeline establishes a new benchmark in auditability and maintainability. Over 92\% of generated code artifacts contain directly traceable commentary, supporting regulatory review and compliance traceability beyond black-box LLM generations. Expert evaluation rates the explanatory alignment of our SS-LLM-SFT outputs substantially higher than prompt-engineered baselines (4.5/5.0 vs. 4.1/5.0).

Modularity in generation enables rapid adaptation to regulatory amendments via localized CU updates, a substantial advantage over sequence-level re-prompting approaches. Error decomposition reveals that 45\% of failures derive from upstream structure prediction-especially relation errors-highlighting targeted priorities for future research in compositional reasoning and semantic modeling. The remaining errors arise from complex logic mappings, entity normalization, and generic code errors, underscoring the hard boundary between language-level and problem-specific model limitations. For more details of error analysis, please refer to Appendix \ref{app:error_study}.

%% file: chapter/7-Conclusion.tex
\section{Conclusion}

We have introduced the \textit{Compliance-to-Code} Dataset and a novel structured synthesis pipeline, embodied in our \textit{FinCheck} system, for executable compliance checking in the complex domain of financial regulation. Our approach, which formalizes regulatory logic as annotated Compliance Units and translates these into verifiable Python code, demonstrates significant gains in functional accuracy (achieving up to 78.1\% Pass@1 with gold structures and reasoning steps), interpretability through traceable commentary, and inherent auditability compared to end-to-end LLM prompting. This research not only provides a valuable new resource and strong benchmark results but also opens new avenues for developing safer, more scalable, and reliable RegTech solutions. 

\textbf{Limitations and Future Work:}
While our approach shows strong results on Chinese financial regulations, extending to other domains and languages may require additional adaptation. Expert annotation remains a bottleneck, though future work could leverage semi-automated or LLM-assisted labeling to improve scalability. Accurately modeling and reasoning about complex legal relationships across documents is still challenging; more advanced graph-based methods or integration of external knowledge bases may help address this. Furthermore, though our current system focuses on translating regulations into compliance logic and granular checks, real-world compliance also involves risk assessment and evaluation of internal controls. Expanding our framework to support these aspects is a promising direction for future research. 

\textbf{Societal Impact:} Our approach can streamline compliance and enhance transparency, but should not replace expert judgment in critical situations. We have carefully reviewed and redacted data for release, and recommend that our tools be used with responsible human oversight. For more discussion, please refer to Ethics Statements in Appendix \ref{app:ethics_statement}.

%% file: chapter/8-License-Accessibility-DataUsage-EthicsStatements.tex
\section{License}
The Compliance-to-Code dataset is publicly available under the
\textbf{Creative Commons Attribution 4.0 International (CC BY 4.0)} license.

The regulatory documents published by the Beijing Stock Exchange (BSE), used to create the dataset, are publicly available.

\section{Accessibility}

1. Links to access the dataset and its metadata. (\url{https://huggingface.co/datasets/GPS-Lab/Compliance-to-Code}). Link to access code (\url{https://github.com/AlexJJJChen/Compliance-to-Code}).

2. The data is saved in both JSON and CSV format, where an example is shown in the README.md file.

3. Logos AI Lab research group will maintain this dataset on the official GitHub account.

4. CC-BY-4.0 (\url{https://github.com/AlexJJJChen/Compliance-to-Code/blob/main/LICENSE}).

\section{Data Usage}
The authors bear all responsibility in case of violation of rights.

\section{Ethics Statements}
\label{app:ethics_statement}

We have considered the broader societal impacts of this work. The development of automated tools for financial compliance checking, such as the proposed \textit{FinCheck} system and the Compliance-to-Code dataset, has the potential for positive societal impact by streamlining compliance processes, increasing transparency, and potentially reducing the costs associated with regulatory adherence. However, we also recognize that such tools should not entirely replace expert human judgment in critical financial and legal situations. Potential negative impacts could arise from over-reliance on automated systems or if the system produces incorrect interpretations, which could lead to erroneous compliance decisions. We recommend that our tools be used with responsible human oversight to mitigate such risks. 

In terms of safeguards for the responsible release of data and models, the dataset derived from public regulatory documents has been carefully reviewed. While the primary risk of misuse for a system designed for regulatory compliance is lower than for some other types of AI models (e.g., generative models for media), we emphasize that the generated code and compliance checks should be validated by domain experts before being relied upon for actual compliance decisions. The data and code will be released with clear documentation regarding its intended use and limitations. 

All existing assets (e.g., pre-trained models or libraries) used in this paper are properly credited, and their licenses and terms of use have been respected. Any new assets introduced, including the Compliance-to-Code dataset and the \textit{FinCheck} pipeline code, will be well-documented, and this documentation will be provided alongside the assets to ensure clarity for users and facilitate reproducibility. 

This research did not involve crowdsourcing or direct research with human subjects; therefore, IRB approval was not applicable. 

The usage of Large Language Models (LLMs) is a core component of the methodology in this research. Specific LLMs employed and their roles in tasks such as regulation structuring and code generation are detailed in the experimental setup section of the paper. This includes information on the base models used and the fine-tuning procedures applied.

%% file: chapter/9-AnnotationSchema.tex
\section{Annotation Schema: Compliance Units and Inter-Unit Relations}
\label{app:cu_annotation}

The foundation of our structured representation is the Compliance Unit, defined as a self-contained, verifiable component of a legal rule. Each Compliance Unit is systematically decomposed into the following constituent fields:

\begin{itemize}
    \item \textbf{Subject:} The entity or set of entities legally bound by the rule (e.g., "controlling shareholder," "directors | supervisors | senior management"). We retain precise legal terminology where appropriate and use "|" to denote a disjunction of subjects.
    \item \textbf{Condition:} The specific context or trigger scenario under which the constraint applies. This field captures behavioral prerequisites, temporal specifications, quantitative thresholds, and states of third parties. Crucially, exception clauses (e.g., "except for judicial enforcement") are integrated logically within the condition, often transformed into negative constraints.
    \item \textbf{Constraint:} The mandatory action, prohibition, or requirement imposed by the rule (e.g., "shall announce 15 days in advance," "must not exceed 3 months," "is prohibited from reducing holdings"). This includes preserving quantitative parameters and explicit modal verbs of obligation ("shall," "must," "is prohibited"). We distinguish between genuine obligations/prohibitions and statements merely conferring rights or permissions; only the former are annotated as constraints unless non-compliance with a permission implies a penalty.
    \item \textbf{Contextual Information:} Ancillary details that are necessary for the correct interpretation or execution of the Compliance Unit but do not fit into Subject, Condition, or Constraint. This includes definitions, calculation methodologies, or references to specific annexes or forms. Content that logically belongs in the core fields is excluded.
\end{itemize}

Regulatory provisions are not usually isolated; they interact logically. We annotate Inter-Unit Relations between Compliance Units to capture these dependencies, enabling the construction of a directed graph representing the legal logic flow. The four primary relation types annotated are:

\begin{itemize}
    \item \textbf{refer to (Reference):} A source Compliance Unit requires information from a target Compliance Unit or an external legal document/definition for complete interpretation or execution. This relation type is primarily utilized during the code generation phase to fetch necessary external information.
    \item \textbf{exclude (Exclusion):} If the conditions of the source Compliance Unit are met, the target Compliance Unit becomes explicitly inapplicable or is overridden. This relation type modifies the outcome during the post-execution evaluation of the compliance graph.
    \item \textbf{only include (Exclusive Applicability):} If the conditions of the source Compliance Unit are met, compliance consideration within the scope of the current regulation is restricted solely to the target Compliance Unit(s), overriding the applicability of other co-located Compliance Units. This relation type also acts during post-execution evaluation.
    \item \textbf{should include (Mandatory Inclusion):} The source Compliance Unit explicitly mandates compliance with the constraint(s) of the target Compliance Unit(s). This relation can imply overriding the original condition(s) of the target unit if the source unit's context directly necessitates the target's constraint (e.g., a general rule (target) vs. a specific scenario (source) that requires the general rule's action regardless of the general rule's specific condition). This relation acts during post-execution evaluation and can influence code logic.
\end{itemize}

%% file: chapter/10-ErrorAnalysis.tex
\section{Error Analysis}
\label{app:error_study}

A detailed analysis of failures in the SS-LLM-SFT (Pred Struct) setting reveals critical insights:
\begin{itemize}
    \item \textit{Upstream Error Propagation (approx. 45\% of errors):} The most significant error source is inaccuracies from Task 1. Key issues include errors in partitioning regulatory text into its Condition and Constraint components, or failing to fully capture and propagate essential contextual information. Erroneous CU boundary detection can also truncate or conflate conditions.
    \item \textit{Complex Logic Mapping Failures (approx. 30\% of errors):} Even with largely correct CUs, mapping highly complex conditional logic, especially involving multiple interacting quantitative thresholds, temporal conditions, or deeply nested exceptions derived from several inter-related CUs, can challenge the code generator. These are typically found in the Difficult problem category.
    \item \textit{Numerical and Entity Precision (approx. 15\% of errors):} Financial regulations demand high precision. Errors include slight miscalculations in formulas derived from text, incorrect mapping of specific named entities (e.g., "controlling shareholder" vs. "actual controller" if not perfectly disambiguated in CUs), or off-by-one errors in date calculations.
    \item \textit{Generic Code Generation Artifacts (approx. 10\% of errors):} These include minor Python syntax errors, incorrect library usage (though rare with code-tuned LLMs), or slight deviations from the exact expected output format for unit tests, not necessarily reflecting a core misunderstanding of the regulation.
\end{itemize}
This detailed error breakdown not only highlights current limitations but also provides a clear roadmap for future research. It emphasizes the need for more robust Task 1 models and advanced compositional generalization in the code generation phase. To effectively handle complex multi-CU scenarios, developing stronger relational reasoning capabilities will also be essential.

%% file: chapter/11-Prompt-of-Instruction-tuning.tex
\section{Prompt of Instruction-tuning}
\label{app:prompt}
\begin{figure*}[h]
    \centering
    \includegraphics[width=1\linewidth]{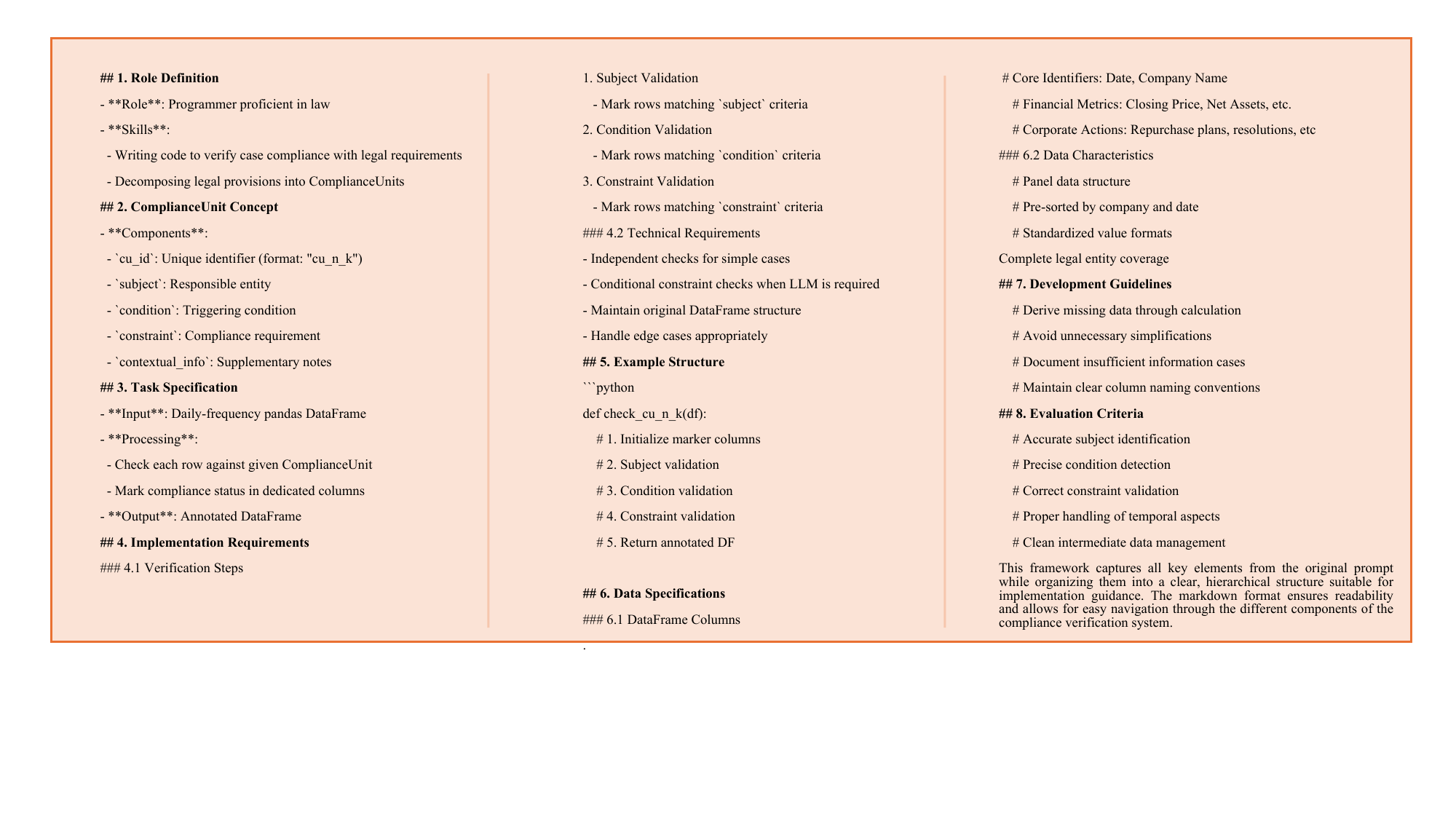}
    \caption{Prompt of Code Generation. }
    \label{fig:prompt}
\end{figure*}

Because of the length of the prompt, we only display the structure of the prompt in Figure \ref{fig:prompt} For more detail, please refer to the dataset.

\begin{table*}[h]
\centering
\small
\caption{Overview of Compliance Control Domains Covered. The columns \# Source clauses, \# Compliance Units, and \# Document Words indicate the number of original regulatory articles, the total count of extracted structured compliance units, and the total word count of the source documents for each domain, respectively.}
\label{tab:regulatory_aspects}
\resizebox{1\textwidth}{!}{%
\begin{tabular}{@{}>{\raggedright\arraybackslash}p{4.5cm} p{7cm} c c c@{}}
\toprule
\textbf{Domain/Aspect} & \textbf{Compliance Control Focus and Risk} & \textbf{\# Source clauses} & \textbf{\# Compliance Units} & \textbf{\# Document Words} \\
\midrule
Independent Director Systems & Independence requirements, director nomination/integrity review, conflict management. & 30 & 126 & 5446 \\
Quarterly Reporting Obligations & Timeliness, content sufficiency, disclosure controls for periodic reports. & 16 & 32 & 1888 \\
Equity Incentives and ESOPs & Plan approval, risk limits, grantee eligibility, compliance disclosure events. & 66 & 154 & 8132 \\
Share Repurchase Controls & Buyback process, threshold triggers, ban periods, reporting duties. & 78 & 266 & 11458 \\
Tender Offer Compliance & Takeover code triggers, procedural controls, reporting/approval checks. & 36 & 113 & 4827 \\
Inside Information Management & Insider identification, ad hoc event control points, information wall requirements. & 25 & 48 & 3564 \\
Board Transfer/Listing Change & Conditions, procedural controls, risk disclosure for inter-market moves. & 21 & 90 & 2990 \\
Large Shareholder Transactions Control & Sale restriction, blackout periods, reporting triggers, conflict discipline. & 29 & 107 & 5504 \\
Raised Fund Use Controls & Escrow, use restriction, monitoring, board/supervisor review. & 27 & 117 & 4623 \\
Equity Rights Distribution & Dividend/bonus triggers, shareholder fairness, process integrity. & 33 & 106 & 4908 \\
\bottomrule
\end{tabular}}
\end{table*}